# Ensemble learning for uncertainty estimation with application to the correction of satellite precipitation products

Georgia Papacharalampous[1], Hristos Tyralis[2,*], Nikolaos Doulamis[3], Anastasios Doulamis[4]

[1] Department of Topography, School of Rural, Surveying and Geoinformatics Engineering, National Technical University of Athens, Iroon Polytechniou 5, 157 80 Zografou, Greece (papacharalampous.georgia@gmail.com, gpapacharalampous@hydro.ntua.gr, https://orcid.org/0000-0001-5446-954X)

[2] Department of Topography, School of Rural, Surveying and Geoinformatics Engineering, National Technical University of Athens, Iroon Polytechniou 5, 157 80 Zografou, Greece (montchrister@gmail.com, hristos@itia.ntua.gr, https://orcid.org/0000-0002-8932-4997)

[3] Department of Topography, School of Rural, Surveying and Geoinformatics Engineering, National Technical University of Athens, Iroon Polytechniou 5, 157 80 Zografou, Greece (ndoulam@cs.ntua.gr, https://orcid.org/0000-0002-4064-8990)

[4] Department of Topography, School of Rural, Surveying and Geoinformatics Engineering, National Technical University of Athens, Iroon Polytechniou 5, 157 80 Zografou, Greece (adoulam@cs.ntua.gr, https://orcid.org/0000-0002-0612-5889)

* Corresponding author

---

---

**Abstract**: Predictions in the form of probability distributions are crucial for effective decision-making. Quantile regression enables such predictions within spatial prediction settings that aim to create improved precipitation datasets by merging remote sensing and gauge data. However, ensemble learning of quantile regression algorithms remains unexplored in this context and, at the same time, it has not been substantially developed so far in the broader machine learning research landscape. Here, we introduce nine quantile-based ensemble learners and address the aforementioned gap in precipitation dataset creation by presenting the first application of these learners to large precipitation datasets. We employed a novel feature engineering strategy, which reduces the number of predictors by using distance-weighted satellite precipitation at relevant locations, combined with location elevation. Our ensemble learners include six that are based on stacking ideas and three simple methods (mean, median, best combiner). Each of them combines the following six individual algorithms: quantile regression (QR), quantile regression forests (QRF), generalized random forests (GRF), gradient boosting machines (GBM), light gradient boosting machines (LightGBM), and quantile regression neural networks (QRNN). These algorithms serve as both base learners and combiners within different ensemble learning methods. We evaluated performance against a reference method (i.e., QR) using quantile scoring functions and a large dataset. The latter comprises 15 years of monthly gauge-measured and satellite precipitation in the contiguous United States (CONUS). Ensemble learning with QR and QRNN yielded the best results across the various investigated quantile levels, which range from 0.025 to 0.975, outperforming the reference method by 3.91% to 8.95%. This demonstrates the potential of ensemble learning to improve probabilistic spatial predictions.

# 1. Introduction

## 1.1 Importance of uncertainty estimation in remote sensing of precipitation

Applications of regression algorithms include those that merge remote sensing and gauge-measured precipitation data (e.g., Baez-Villanueva et al. 2020; Nguyen et al. 2021; Sui et al. 2022; Papacharalampous et al. 2023a). These are often termed “spatial interpolation”, “bias correction” or “satellite product blending”, depending on context. Still, most of them predominantly fall within the general “spatial prediction” category (Hengl et al. 2018).

Merging remote sensing and gauge-measured precipitation data with machine learning is recognised as an important endeavour in earth observation and geoinformation (Hu et al. 2019; Abdollahipour et al. 2022), as it can lead to spatially dense datasets with larger accuracy than the remote sensing ones. The various remote sensing machine learning regression applications usually issue point predictions through machine learning algorithms such as those described in Hastie et al. (2009), James et al. (2013) and Efron and Hastie (2016). These predictions provide a small amount of information.

The notion of uncertainty estimation in such settings refers to the requirement that precipitation predictions, at every point in space, be given in the form of a probability distribution, instead of a point prediction. Precipitation is a continuous variable; therefore, such problems are regression problems, not classification problems. Uncertainty estimation is essential because of the large amount of information that it provides to decision makers (Gneiting and Raftery 2007).

## 1.2 Existing literature

The literature devoted to methods for uncertainty estimation with machine learning in the remote sensing of precipitation is limited, with representative examples being demonstrated by Bhuiyan et al. (2018), Zhang et al. (2022), Glawion et al. (2023), Tyralis et al. (2023) and Papacharalampous et al. (2024; 2025). This holds although machine learning offers several advantages such as improved predictive performance and convenient implementation (Papacharalampous and Tyralis 2022; Tyralis and Papacharalampous 2024). All the existing relevant approaches are based on individual algorithms, with none of the available studies having explored ensemble learning for uncertainty estimation (excluding combinations of distributional regression algorithms developed by Papacharalampous et al. 2025), despite its well-established performance improvement over individual algorithms in point prediction applications (Sagi and Rokach 2018; Papacharalampous and Tyralis 2022; Wang et al. 2023; Tyralis and Papacharalampous 2024).

Ensemble learning can be made in either simple (e.g., the hard to beat in practice equal weight averaging proposed, for instance, by Smith and Wallis 2009; Petropoulos and Svetunkov 2020; Lichtendahl et al. 2023) or complex (e.g., Wolpert 1992) ways. Methods for combining probabilistic predictions also include Bayesian model averaging and non-linear pooling, among others (Wang et al. 2023). However, to the best of our knowledge, non-linear combinations of probabilistic predictions based on machine learning-based combiners are introduced here for the first time.

## 1.3 Aims of the study and methodological contributions

Essentially, the problem we aim to solve is that of improving satellite precipitation products based on gauge-measured data. The term "improving" here refers to the application of machine learning to predict precipitation at every point in space using satellite data as predictors. A distinctive attribute of our study is that its predictions are probabilistic (in the form of multiple quantiles) instead of point predictions.

The aims of the study and its methodological contributions are outlined as follows:

**a. Methodological contributions in machine learning**

We introduce and evaluate non-linear ensemble learning of probabilistic predictions in the form of quantiles. By minimising quantile loss functions (Gneiting 2011), we specifically developed ensemble learning methods for this purpose. To this end, individual algorithms and their combiners are machine learning quantile regression algorithms (such as linear quantile regression (QR), quantile regression forests (QRF), quantile-based boosting and quantile regression neural networks (QRNN)). The methods are evaluated in predicting quantiles of the predictive probability distribution at multiple levels using quantile loss functions. We deviate from previously published methods for combining algorithms for estimating uncertainty, since we introduce combinations of quantile regression algorithms instead of distributional regressions (demonstrated by Papacharalampous et al. 2025), in line with our focus on improved performance. In particular, quantile regression is more flexible compared to distributional regression, thus allowing for better predictions in practical situations (Tyralis and Papacharalampous 2024).

**b. Contribution to remote sensing of precipitation**

Our methods improve probabilistic predictions of precipitation in spatial settings compared to existing methods (see Section 1.2), particularly when merging remote sensing and gauge-measured precipitation data. The magnitude of the improvements was quantified here using quantile loss functions. Consequently, the proposed methods are potentially useful in both spatial statistics and remote sensing.

The quantile-based ensemble learning methods demonstrated significant improvement over previous approaches, which are based on individual algorithms, in the topic of uncertainty estimation in the remote sensing of precipitation.

### 1.4 Application and paper outline

We applied the methods to a dataset comprising 15-year-long monthly gauge-measured and satellite precipitation data from across the Contiguous United States (CONUS). The gauge measurements served as the ground truth, while the satellite data and gauge elevation played the role of predictor variables. We employed a feature engineering method that halves the number of predictor variables by using distance-weighted satellite data, instead of raw satellite data and distances. This approach eliminates redundant predictor values, particularly distances, potentially improving the performance of non-tree-based algorithms (e.g., Papacharalampous et al. 2023a, b, c, 2024; Tyralis et al. 2023).

The remaining article is structured in five sections. Section 2 describes the ensemble learning methods and their elements, while Section 3 describes how these methods where applied in this work, including the new feature engineering strategy in spatial prediction, and the data used for this application. Section 4 presents the results, which are then discussed in view of the existing literature in Section 5. Section 6 concludes the article.

## 2. Ensemble learners

### 2.1 Base learners

Uncertainty estimation can be conducted through a variety of machine learning algorithms (Tyralis and Papacharalampous 2024). Herein, the interest was in ensemble learners that predict the quantile of the probability distribution of continuous random variables. Therefore, the problem can be formulated in a regression setting. For constructing such ensemble learners, we used algorithms from the quantile regression family (see Table 1). Such algorithms either optimise across a training dataset the quantile scoring function (e.g. QR, gradient boosting machines (GBM), light gradient boosting machines (LightGBM) and QRNN), a scoring function that is strictly consistent for the quantile of the probability distribution (Gneiting 2011), or have been proven optimal for predicting a quantile (QRF and generalized random forests (GRF)). The property of strict consistency of quantile scoring functions for the quantile incentivizes modellers to be honest when evaluating their quantile predictions. This holds in the sense that, when one receives a directive to predict a quantile, the expected quantile loss is minimised when following the directive (Gneiting 2011). The quantile scoring function is defined, as in

$$L_\tau(z, y) := (z - y)(\mathbb{I}(z \geq y) - \tau), \quad (1)$$

where $\tau$, $y$ and $z$ are the quantile level, the observation and the prediction, respectively, and $\mathbb{I}(A)$ is the indicator function, which is equal to 1 when the event $A$ realises and equal to 0 otherwise. Further information on the quantile scoring function relevant to remote sensing applications can be found in Papacharalampous et al. (2024).

Table 1. Individual algorithms used for forming each of the ensemble learners.

| Name | Abbreviation | Reference(s) |
|---|---|---|
| Quantile regression | QR | Koenker and Bassett (1978); Koenker (2005) |
| Quantile regression forests | QRF | Meinshausen and Ridgeway (2006) |
| Generalized random forests | GRF | Athey et al. (2019) |
| Gradient boosting machines | GBM | Friedman (2001); Mayr et al. (2014) |
| Light gradient boosting machines | LightGBM | Ke et al. (2017) |
| Quantile regression neural networks | QRNN | Taylor (2000); Cannon (2011) |

## 2.2 Ensemble learners

Let us suppose that we are interested in predicting the quantile at level $\tau$ by combining the independent predictions of two or more quantile regression algorithms (base learners). For this case, we propose the utilisation of a quantile regression algorithm as the combiner. Under this strategy, the predictions of the base learners for the quantile at level $\tau$ are used as predictor variables for the combiner, with the predictand being the quantile at level $\tau$. Due to the properties of the quantile regression algorithms (minimisation of the quantile scoring function), the strategy introduced is expected to lead to optimal predictions for the quantile compared to the base learners (van der Laan 2007; Wolpert 1992; Yao et al. 2018). A pseudo algorithm for the implementation of the ensemble in a training set of $n$ samples follows:

- Step 1: Split the training set randomly into set 1 with $n_1$ samples and set 2 with $n_2$ samples (in our application, we set $n_1 = n_2$ (see Section 3.2), though the existing literature is inconclusive regarding the optimal split ratio), where $n_1 + n_2 = n$.
- Step 2: Train quantile regression algorithms $1, \dots, k$ in set 1 to predict quantile $q_\tau$ at level $\tau$. Let the predictions of the set 1 trained algorithms in set 2 be notated with $q_{1,\tau}, \dots, q_{k,\tau}$ respectively.
- Step 3: Train the combiner in set 2 using $q_{1,\tau}, \dots, q_{k,\tau}$ as predictors to minimise the quantile average score at level $\tau$.

- Step 4: Retrain quantile regression algorithms $1, \dots, k$ in the full training set (union of set 1 and set 2) to predict quantiles $q_\tau$ at level $\tau$. Let the predictions of the full set trained algorithms in the test set be notated with $q_{\mathrm{upd},1,\tau}, \dots, q_{\mathrm{upd},k,\tau}$ respectively.
- Step 5: Issue quantile predictions in the test set with the trained combiner of step 3 using $q_{\mathrm{upd},1,\tau}, \dots, q_{\mathrm{upd},k,\tau}$ as predictor values.

Six ensemble learning methods were formulated based on the above concepts. All of them use the total of the individual quantile regression algorithms in Table 1 as their base learners and each of them uses a different individual quantile regression algorithm as its combiner. The parameters of the individual algorithms are set as in Papacharalampous et al. (2024). To provide benchmarks for these ensemble learners (other than the individual quantile regression algorithms in Table 1, which are also reasonable benchmarks), three simple ensemble learners were formulated. Their base learners are the same as for the ensemble learning methods that are based on stacking ideas, and their combiners are the mean of the predictive quantiles, the median of the predictive quantiles and the best learner. The ensemble learners are outlined in Figure 1. In summary, 15 algorithms were compared, including the six base learners, the six ensemble learning algorithms that are based on stacking ideas, the mean combiner, the median combiner and the best learner.

**How does each ensemble learner work?**

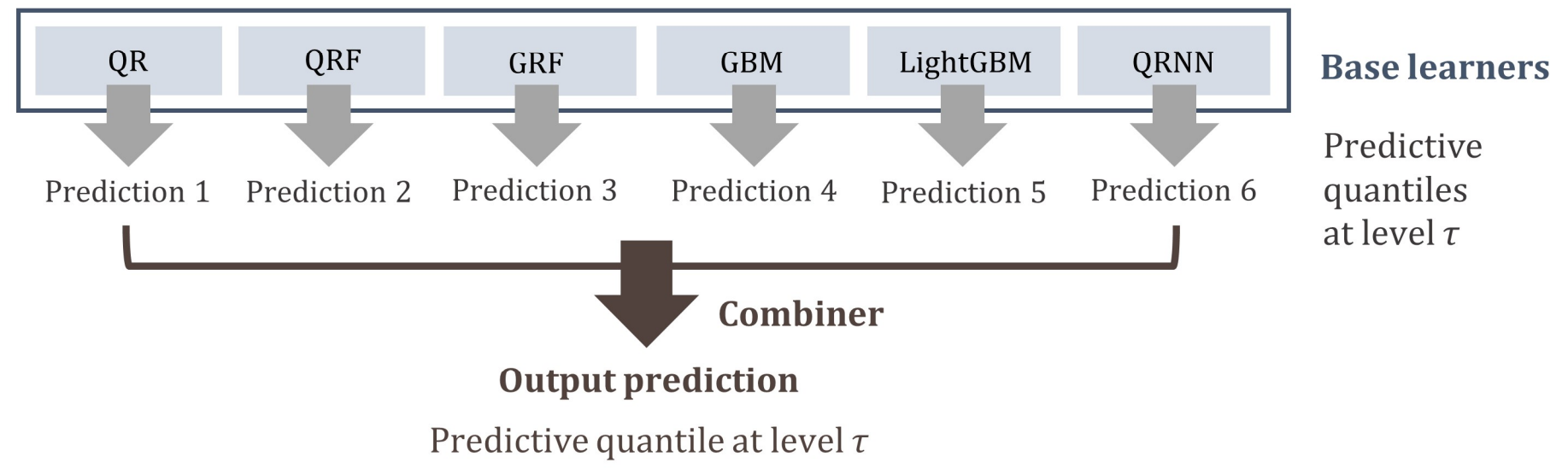

**What is the difference between the ensemble learners?**

They utilize different combiners.

**Which are the combiners introduced in this study?**

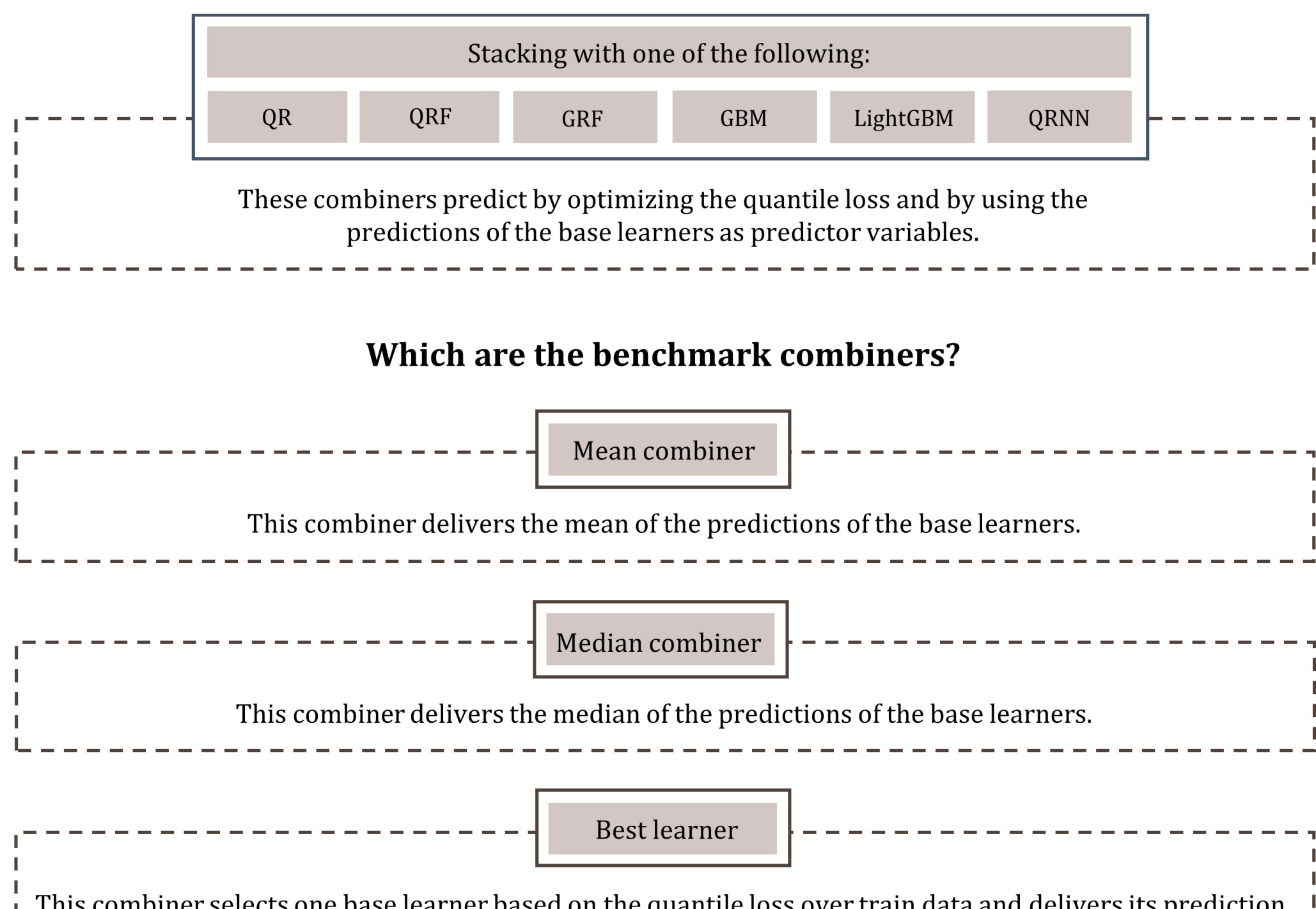

Figure 1. Ensemble learners formulated in this study and their combiners. QR, QRF, GRF, GBM, LightGBM and QRNN stand for quantile regression, quantile regression forests, generalized random forests, gradient boosting machines, light gradient boosting machines and quantile regression neural networks, respectively.

## 2.3 Plain language summary

To summarise the theory presented in Sections 2.1 and 2.2 in plain language, we focus on the following four points:

### *2.3.1 How simple quantile regression algorithms issue probabilistic predictions*

The core idea of quantile regression is to train a regression algorithm using the quantile scoring function (1). This scoring function is strictly consistent for evaluating quantile predictions, as discussed in Section 2.1. Consequently, during training, the same function also acts as a consistent estimator for regression models, as demonstrated by Dimitriadis et al. (2024).

This has significant implications, in the sense that the type of scoring function used in training directly determines the predictions issued by the model. For example, a model trained with a squared error scoring function will predict the mean of the conditional distribution of the dependent variable.

By contrast, training a regression model with a quantile scoring function at level $\tau$ enables it to predict conditional quantiles at that level. Repeating this process across a dense grid of quantile levels produces an approximation of the conditional distribution of the prediction, effectively yielding an uncertainty estimate for the prediction.

*2.3.2 When probabilistic predictions are considered reliable in absolute terms.*

Suppose a model issues a quantile prediction at a pre-specified level $\tau$. The question arises: When is this prediction reliable in absolute terms? To address this, we define the $\tau$-quantile identification function $V_\tau$ (Gneiting 2011)

$$V_\tau(z, y) = \mathbb{1}(z \geq y) - \tau \quad (2)$$

where $\tau$, $y$ and $z$ are the quantile level, the observation and the prediction, respectively. The function $V_\tau$ identifies quantiles in the sense that, if a prediction $Q^\tau$ corresponds to the true $\tau$-quantile of a probability distribution $F$, then the expectation $\mathbb{E}_F[V_\tau(Q^\tau, \underline{y})]$ equals zero, where $\underline{y}$ is a random variable following the distribution $F$.

The identifiability property of the quantile is critical because it enables assessing the absolute reliability of predictions based on quantile regression algorithms (Fissler et al. 2021). This reliability assessment aligns with the statistical concept of calibration, as outlined by Fissler et al. (2023) and Gneiting and Resin (2023).

In practical settings, we can assess whether predictions from a quantile regression algorithm are reliable in absolute terms only if the algorithm generates predictions at multiple quantile levels. Let $k$ be the number of samples in the test set, and $y_i$ and $z_i$, $i \in \{1, \dots, k\}$ are the observation and $\tau$-quantile prediction at each point $i$ of the test set. The predictions are deemed reliable if the coverage

$$\text{coverage} = (1/k) \sum_{i=1}^{k} V_\tau(z_i, y_i) + \tau \quad (3)$$

equals to the nominal quantile level $\tau$. Equivalently this occurs when $(1/k)\sum_{i=1}^{k} V_\tau(z_i, y_i) = 0$.

Intuitively, the term $(1/k)\sum_{i=1}^{k} \mathbb{1}((z_i \geq y_i)$ calculates the proportion of predictions that exceed the observed values. For predictions to be reliable, this proportion should match the nominal quantile level $\tau$. For example, if $\tau = 0.9$, we expect 90% of predictions to exceed the actual observations.

*2.3.3 How to compare two probabilistic predictions.*

In practice, predictions are often reliable, necessitating a method to rank their performance. Scoring functions, such as the quantile scoring function, address this need by enabling the comparison and ranking of predictions (Fissler and Ziegel 2016). To build intuition, consider a regression algorithm that issues quantile predictions at two levels $\tau_1$ and $\tau_2$, where $\tau_1 < \tau_2$. Let $(a_1, a_2)$ denote the resulting prediction interval for the first algorithm and $(b_1, b_2)$ for a second algorithm. While both intervals may be reliable, we typically prefer the sharper interval (i.e., the narrower one). For instance, if $a_2 - a_1 < b_2 - b_1$, the first interval is sharper. A strictly consistent scoring function for prediction intervals would rank the first algorithm higher. The same logic applies to quantile predictions, though the intuition is less direct.

*2.3.4 Ensemble learning for improving probabilistic predictions*

Based on the earlier discussion, the proposal for non-linear ensemble learning of quantile regression algorithms is motivated by two main arguments:

a. Previous remote sensing studies have tested individual algorithms for issuing probabilistic predictions, which were shown to be reliable (Papacharalampous et al. 2024). The natural next step is to improve these algorithms using ensemble learning, which directly minimises the strictly consistent scoring function used for evaluation. This scoring function is essential because it ranks competing predictions. As discussed in Section 1.2, ensemble learning is likely to improve the performance of individual models.

b. Non-linear ensembles that use machine learning combiners (which have not been studied in prior work, see Section 1.2) could outperform linear combiners by better modelling complex relationships between base models. Testing these non-linear ensembles in our specific application is recommended.

# 3. Datasets and application

## 3.1 Datasets

We applied the nine ensemble learners (see Section 2) and the six individual algorithms (QR, QRF, GRF, GBM, LightGBM and QRNN) for estimating uncertainty while merging remote sensing and gauge-measured data. For this application, data from four databases were sourced (see Table 2). The same data were previously exploited all together in different experiments by Papacharalampous et al. (2023c, 2024).

Table 2. Databases from which data were retrieved for this study and data retrieval details.

| Dataset | Name | Source | Address | Date accessed | Data type | Reference |
|---|---|---|---|---|---|---|
| GHCNm | Global Historical Climatology Network monthly database, version 2 | National Oceanic and Atmospheric Administration (NOAA) | https://www.ncei.noaa.gov/pub/data/ghcn/v2 | 2022-09-24 | Gauge-measured precipitation | Peterson and Vose (1997) |
| PERSIANN | Precipitation Estimation from Remotely Sensed Information using Artificial Neural Networks | Centre for Hydrometeorology and Remote Sensing (CHRS), University of California, Irvine (UCI) | https://chrsdata.eng.uci.edu | 2022-03-07 | Remote sensed precipitation | Hsu et al. (1997); Nguyen et al. (2018); Nguyen et al. (2019) |
| IMERG | GPM Integrated Multi-satellitE Retrievals late precipitation L3 1 day 0.1 degree x 0.1 degree V06 | National Aeronautics and Space Administration (NASA) Goddard Earth Sciences (GES) Data and Information Services Center (DISC) | https://doi.org/10.5067/GPM/IMERGDL/DAY/06 | 2022-12-10 | Remote sensed precipitation | Huffman et al. (2019) |
| AWSTT | Amazon Web Services Terrain Tiles | Amazon Web Services (AWS) | https://registry.opendata.aws/terrain-tiles | 2022-09-25 | Elevation | – |

The precipitation data refer to the years 2001–2015 and to the locations shown in Figures 2 and 3 for the gauge-measured and the remote sensing data, respectively. In particular, 1 421 gauges offered data for this study, while the spatial resolution of both remote sensing datasets is 0.25 degree × 0.25 degree. Bilinear interpolation was applied to the original IMERG dataset for obtaining precipitation at this spatial resolution, as the PERSIANN dataset was available at it. Furthermore, as the PERSIANN and IMERG data originally extracted were daily in opposition to the total monthly data from GHCNm, total monthly PERSIANN and IMERG data had to be formed through time series aggregation. The elevation data refer to the locations shown in Figure 2 (i.e., the locations of the ground-based stations).

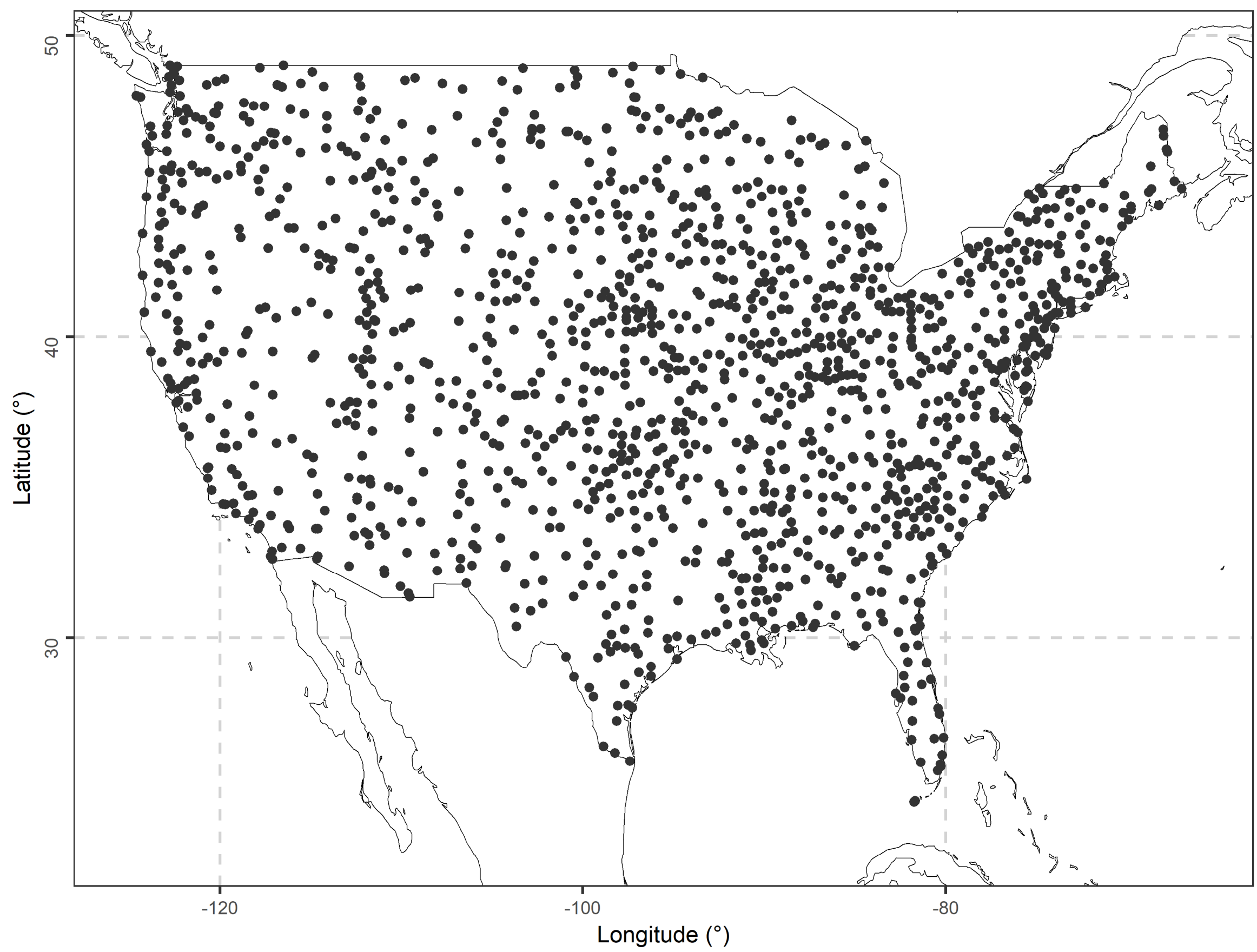

Figure 2. Locations of the ground-based stations that offered time series for this study.

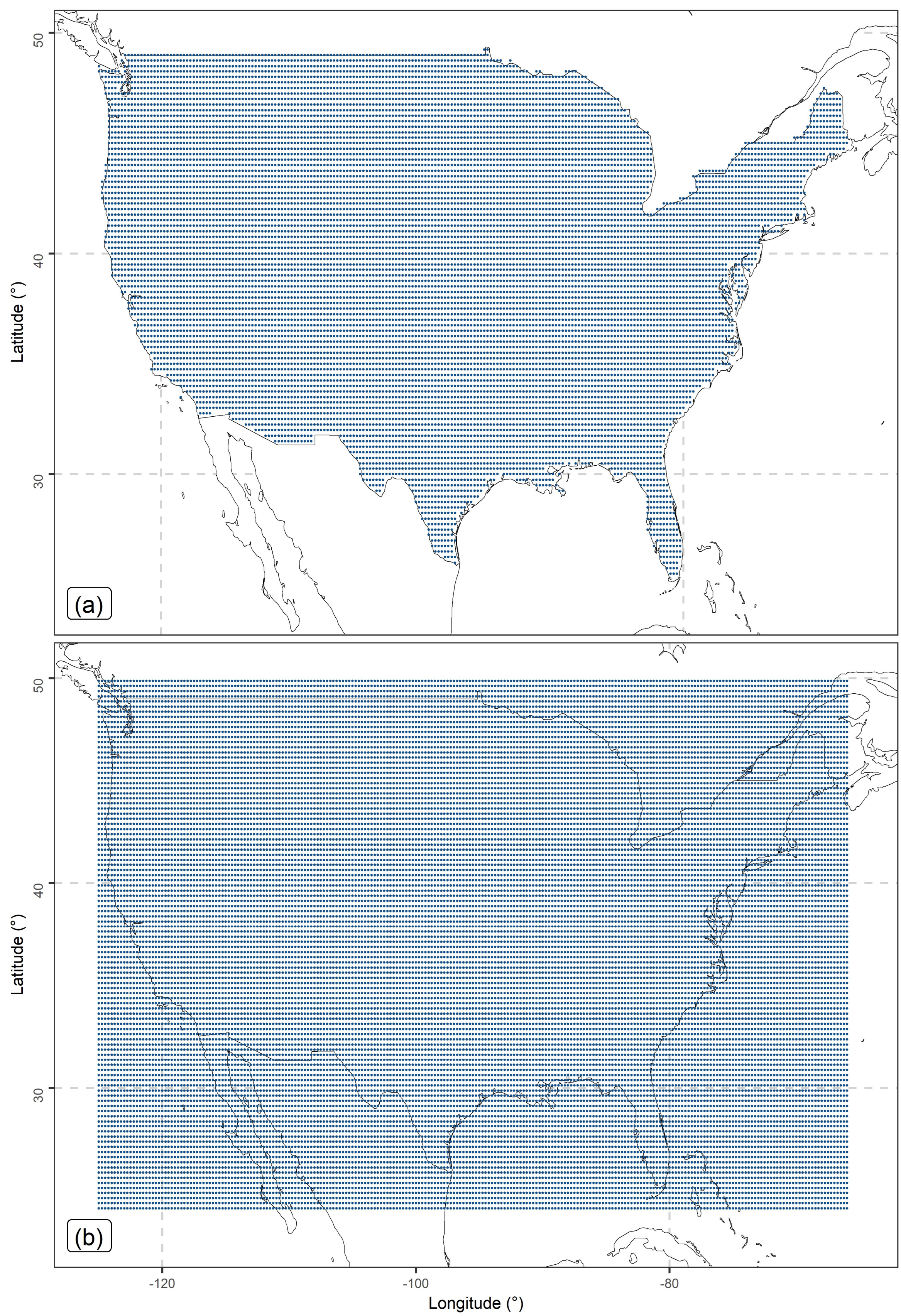

Figure 3. Centre locations of the (a) PERSIANN and (b) IMERG data grids.

## 3.2 Algorithm implementation

The remote sensing data are inaccurate but available at a dense spatial grid. On the other hand, the gauge-measured data are accurate but available only for the locations shown in Figure 2. To form accurate precipitation data at a dense spatial grid via prediction, we can merge remote sensing and gauge-measured precipitation data using machine learning. To simultaneously assess the uncertainty of the new data and, therefore, provide probabilistic instead of point predictions, we can use machine learning algorithms such as those described in Section 2. For the merging, the remote sensing data can be used as predictor variables, together with topography variables, and the gauge-measured data should take the role of the predictand because they are the ground-truth. Relevant spatial prediction settings are available, for instance, in Baez-Villanueva et al. (2020) and Papacharalampous et al. (2023b).

Let the distances of a given station (station 1) from its four closest grid points (grid points 1−4, where grid points refer to the centre of the grid) be denoted with $d_i$, where $i = 1, 2, 3$ and 4 (Figure 4). Herein, these distances and the remote sensing data at the same grid points were used to apply distance-based weighting, separately for each remote sensing dataset. More precisely, the distance-weighted precipitation $\dot{\mathrm{PR}}_k$ at grid point $k = 1, \dots, 4$, is defined, as in

$$\dot{\mathrm{PR}}_k := \frac{(1/d_k^2)\mathrm{PR}_k}{\sum_{i=1}^{4} 1/d_i^2}, k = 1, \dots, 4, \tag{4}$$

where $\mathrm{PR}_k$ is the raw satellite precipitation at grid point $k$. The variables represented by the distance-based weighted precipitation values are referred to hereinafter as “PERSIANN variables 1−4” and “IMERG variables 1−4”, and are the predictor variables for predicting the precipitation value at station 1, together with the elevation at the same station. Using distance-weighted precipitation allows us to halve the number of predictor variables (compared to using raw precipitation data and distances as predictors) in a physically principled manner (closer points are assigned higher weights), while simultaneously reducing the redundancy introduced by repeated distance values. However, while the reduced number of variables may contain less information, this is potentially compensated for by the potentially improved performance of non-tree-based algorithms, which might be sensitive to such redundancies.

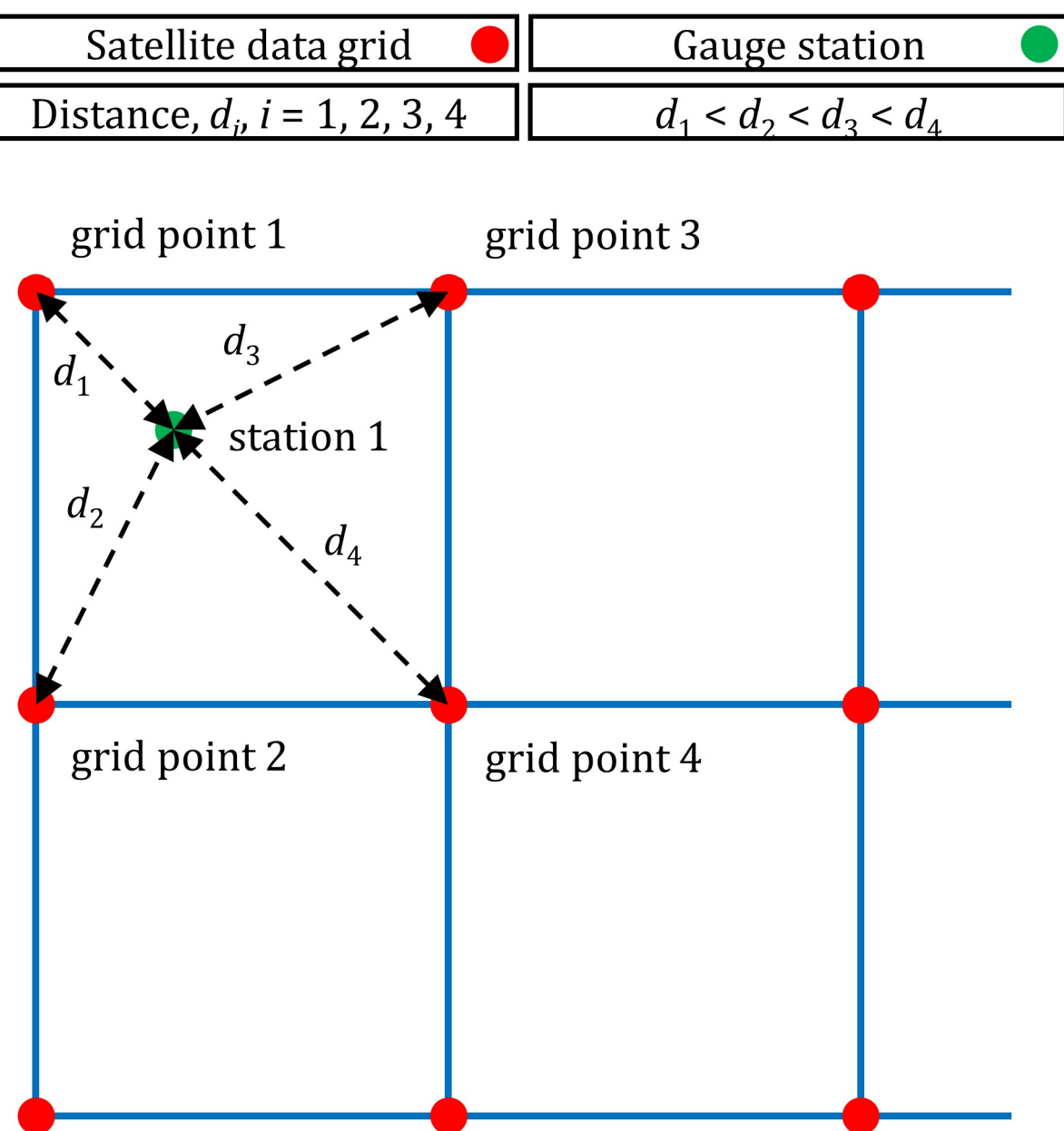

Figure 4. Technical details of the application of the algorithms in this study. The remote sensing data are inaccurate but available at all the grid points (grid points refer to the centre of the grid), while the station-measured data are accurate but available only for the locations shown in Figure 2. Precipitation at the stations is the target variable. The distances of a given station (station 1) from its four closest grid points (grid points 1−4) are denoted with $d_i, i = 1, 2, 3$ and 4. These distances and the remote sensing precipitation data at the same grid points were used to compute eight predictor variables with distance-based weighting.

The dataset was composed by 91 623 samples, each of which contained 10 values. In particular, a sample is of the form sample$_i$ = {$\mathrm{PR}_{\mathrm{station}}$, $\dot{\mathrm{PR}}_{1,\mathrm{IMERG}}$, $\dot{\mathrm{PR}}_{2,\mathrm{IMERG}}$, $\dot{\mathrm{PR}}_{3,\mathrm{IMERG}}$, $\dot{\mathrm{PR}}_{4,\mathrm{IMERG}}$, $\dot{\mathrm{PR}}_{1,\mathrm{PERSIANN}}$, $\dot{\mathrm{PR}}_{2,\mathrm{PERSIANN}}$, $\dot{\mathrm{PR}}_{3,\mathrm{PERSIANN}}$, $\dot{\mathrm{PR}}_{4,\mathrm{PERSIANN}}$, $\mathrm{elevation}_{\mathrm{station}}$}, $i =$ 1, … ,91 623, where $\mathrm{PR}_{\mathrm{station}}$ is the observed precipitation at a station in a specified month, $\dot{\mathrm{PR}}_{k,\mathrm{IMERG}}$ and $\dot{\mathrm{PR}}_{k,\mathrm{PERSIANN}}$, $k = 1, \dots ,4$ are the distance-weighted satellite precipitations in the same month and $\mathrm{elevation}_{\mathrm{station}}$ is the station's elevation. In the regression setting, $\mathrm{PR}_{\mathrm{station}}$ is the dependent variable and the sample's remaining variables are the predictors.

The dataset was randomly split into three equally-sized sets. The first of these sets was used to train the individual algorithms (which were applied as described in Papacharalampous et al. 2024), and the second for making predictions of the same algorithms. The predictions for the second set were used by the best learner, together with their corresponding true values, for identifying a single best algorithm based on the quantile scoring function averaged across the samples of set 2. They were also used as predictor variables by the ensemble learning algorithms for training the combiners to predict the true values (see the pseudo algorithm in Section 2.2). Then, the individual algorithms were trained on the union of sets 1 and 2, and predictions were obtained for set 3. These predictions were used for forming the predictions of all the ensemble learners for set 3 (see the pseudo algorithm in Section 2.2). Additionally, they were used for benchmarking the ensemble learners.

We note that the setting of the prediction problem allows for the following:

a. There is no need to fill missing values in the gauge-measured data. Such filling introduces uncertainties that one seeks to avoid. In particular, if a sample includes a missing value, it is simply discarded; however, remaining samples at the same time but at a different location are kept, as one can compute their predictors.

b. The algorithms, once trained, can predict precipitation at any point in space, because predictors are always available.

c. There is no need to discard stations, even if multiple of them fall inside a square, as depicted in Figure 4. That is because they play the role of different samples, with the values of the predictors differing at each sample.

Predictive quantiles at a dense grid consist an approximation of the predictive probability distribution. In this work, predictions were made for the quantile levels $\tau \in \{0.025, 0.050, 0.075, 0.100, 0.200, 0.300, 0.400, 0.500, 0.600, 0.700, 0.800, 0.900, 0.925, 0.950, 0.975\}$. As precipitation cannot be negative, negative predictions at the quantile were set to zero. To ensure that predictive quantiles do not cross, for each set {data sample, algorithm}, any prediction that was smaller than the prediction of the immediate lower quantile level was set equal to the latter prediction.

### 3.3 Performance comparison

For each set {predictive $\tau$-quantile, algorithm}, a quantile score was computed according to Equation (1) in the test set. Then, separately for each algorithm, the quantile scores were averaged over the test dataset, as in

$$\bar{L}_\tau(z, y) := (1/k) \sum_{i=1}^{k} L_\tau(z_i, y_i), \tag{5}$$

where $k$ is the number of samples included in the test dataset, and $y_i$ and $z_i$, $i \in \{1, \dots, k\}$ are the observation and $\tau$-quantile prediction, respectively, of the $i^{\text{th}}$ sample.

As the average quantile scores, are not scaled, quantile skill scores were computed, as in

$$\bar{L}_{\tau,\text{skill}} := 1 - \bar{L}_{\tau,\text{algorithm}} / \bar{L}_{\tau,\text{benchmark}}, \quad (6)$$

where the benchmark is QR, which is the simpler algorithm. The quantile skill score takes values between $-\infty$ and 1. Quantile skill score larger (smaller) than zero indicates that the predictions of the algorithm of interest are better (worse) than the predictions of the benchmark. Quantile skill score equal to 1 indicates that the predictions of the algorithm of interest are perfect. For an easier comparison between the algorithms, their ranking based on the quantile skill score was obtained for each quantile level.

Additionally, frequencies (sample coverages) were computed. More precisely, for each set {algorithm, quantile level} and for the entire dataset, the frequency with which the prediction is smaller or equal to its corresponding observation was computed. The closer the sample coverages to their nominal values, the larger the reliability of the predictions.

### 3.4 Predictor variable importance

GRF and LightGBM were additionally used to investigate predictor variable importance in two settings. In the first one, the predictor variables were the IMERG variables 1−4, the PERSIANN variables 1−4 and the elevation at the station when GRF and LightGBM were trained on the entire data sample. In the second setting, the predictor variables were the predictions by the base learners (QR, QRF, GRF, GBM, LightGBM and QRNN) in the ensemble learning frameworks having GRF and LightGBM as their combiners. For each set {setting, quantile level, predictor variable}, a simple weighted sum of how many times the predictor variable was split on at each depth in the forest was computed through GRF (Tibshirani and Athey 2023), and the total gain in splits (Shi et al. 2023) was computed through LightGBM. These statistics should be interpreted as follows: The larger their values, the larger the importance of the predictor variable. Based on this, ranks of the predictor variables at each quantile level were obtained. The smaller the rank of a predictor variable, the more important this predictor variable.

## 4. Results

### 4.1 Comparison of algorithms

In summary, the algorithms predicted quantiles at several levels. Thus, their comparison should rely on a scoring function that is strictly consistent for the quantile. Herein, we selected the quantile scoring function. To facilitate comparisons across the entire sample, we computed quantile skill scores. The latter are presented in Figure 5a, while the ranks of the algorithms based on these scores are presented in Figure 5b. For all the quantile levels, ensemble learning with QR and ensemble learning with QRNN are the two best-performing algorithms. For the quantile levels {0.075, 0.100, 0.200, 0.300, 0.400, 0.500, 0.600, 0.700, 0.800}, LightGBM and the best learner exhibit the same performance and share the third position. Other algorithms that exhibit good performance are ensemble learning with GBM from the ensemble learners, and QRF and GBM from the individual algorithms. The mean and median combiners are ranked before QR, GBM and QRNN, but after the remaining individual algorithms. The worst among all the ensemble learning methods, for the problem investigated, is ensemble learning with LightGBM.

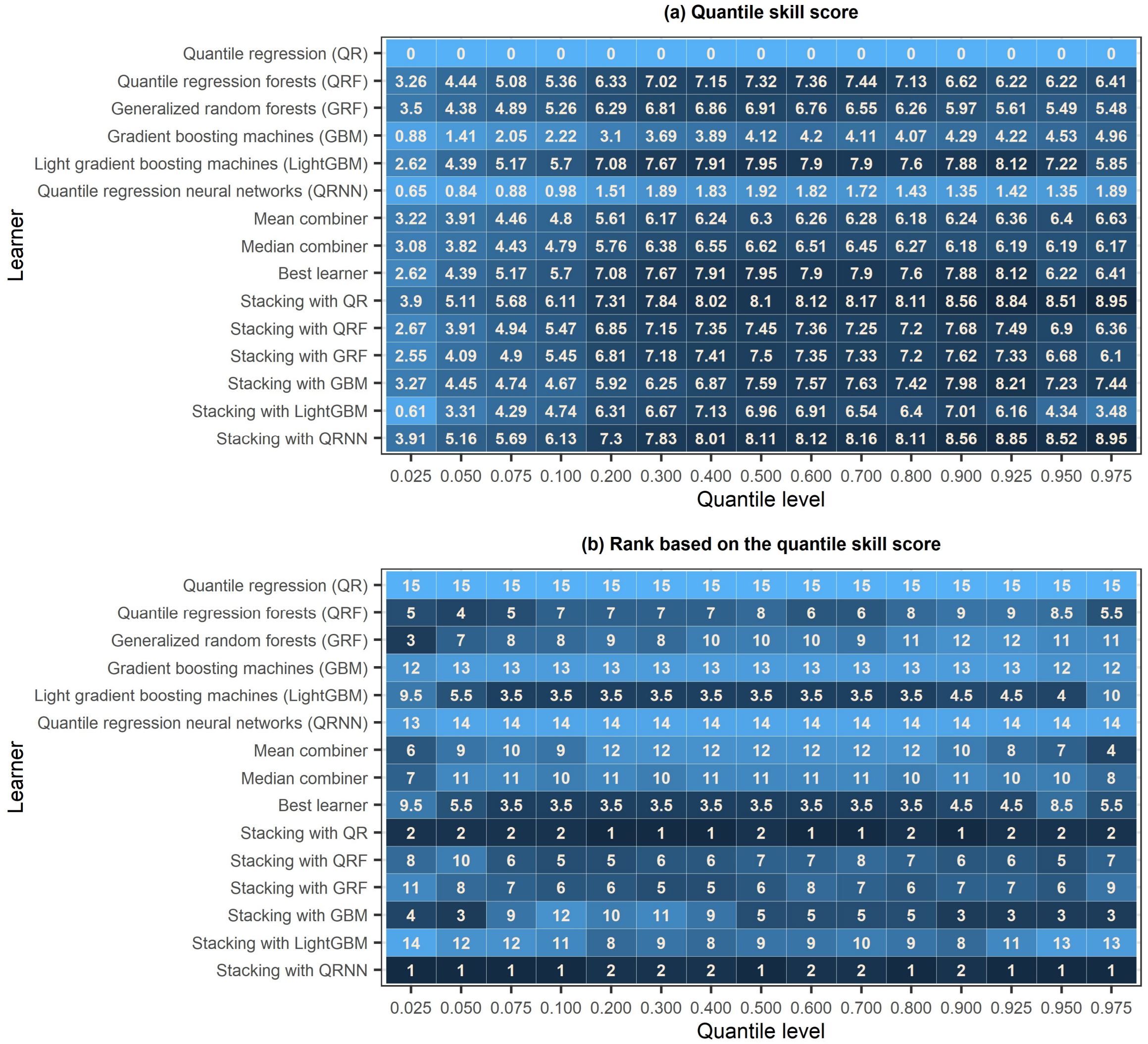

Figure 5. (a) Quantile skill score and (b) rank of each of the algorithms at the various quantile levels. The larger the quantile skill score, the smaller the rank and the darker the colour, the better the predictions on average compared to the predictions of quantile regression.

The sample coverages of the quantile predictions at the various quantile levels could also be of interest. Figure 6 shows that these statistics are close to their nominal values for the quantile predictions of all the algorithms. Although predictive coverages are intuitive and help us to understand whether the predictions are good in an absolute sense, they are not consistent (please recall the definition of consistency of a scoring function in Section 2.1). To this end, ranking of the algorithms should be based on quantile scoring functions, as presented in Figure 5. Recall from Section 2.1, that quantile scoring functions are consistent for quantiles; therefore, they encourage the assessor to be honest when evaluating quantile predictions (Gneiting 2011).

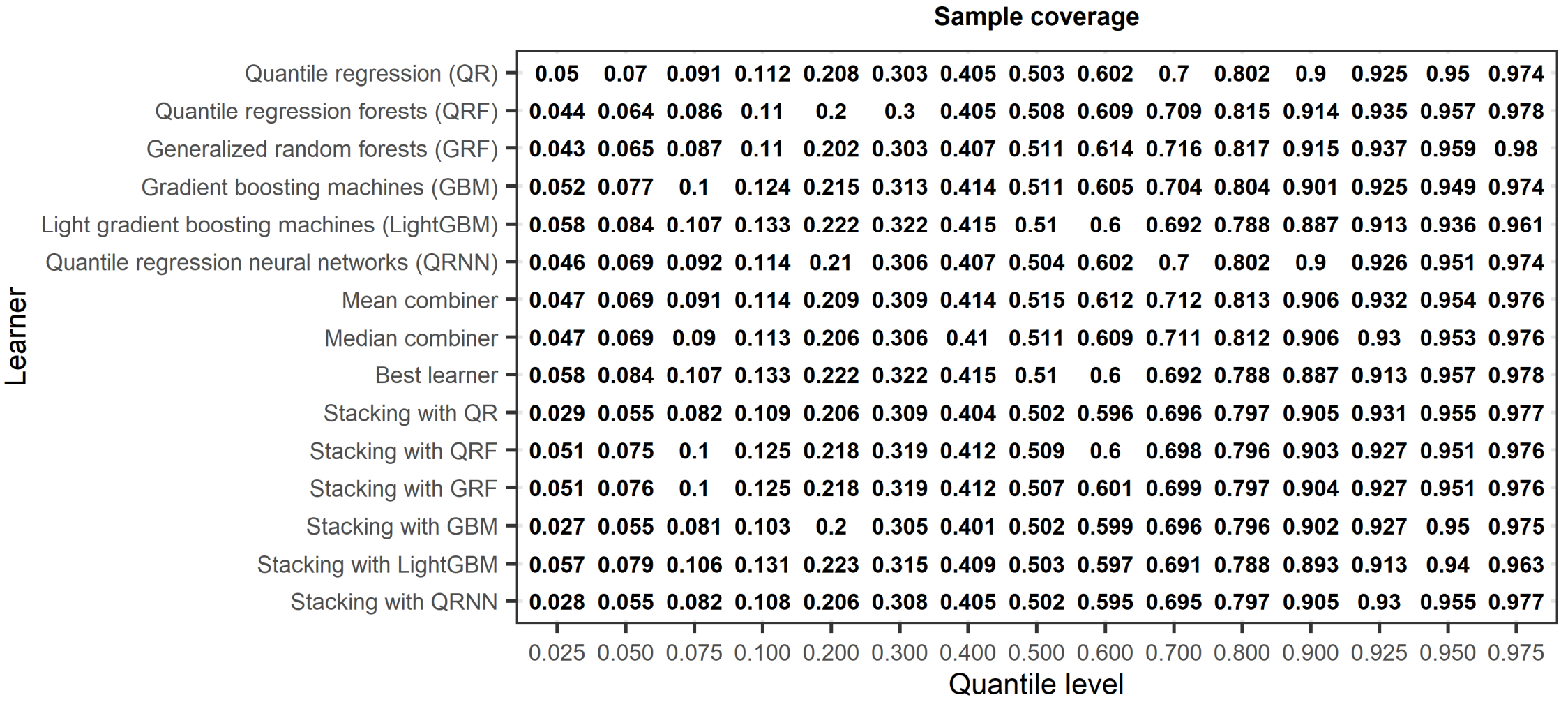

Sample coverage

| Learner / Quantile level | 0.025 | 0.050 | 0.075 | 0.100 | 0.200 | 0.300 | 0.400 | 0.500 | 0.600 | 0.700 | 0.800 | 0.900 | 0.925 | 0.950 | 0.975 |
|---|---|---|---|---|---|---|---|---|---|---|---|---|---|---|---|
| Quantile regression (QR) | 0.05 | 0.07 | 0.091 | 0.112 | 0.208 | 0.303 | 0.405 | 0.503 | 0.602 | 0.7 | 0.802 | 0.9 | 0.925 | 0.95 | 0.974 |
| Quantile regression forests (QRF) | 0.044 | 0.064 | 0.086 | 0.11 | 0.2 | 0.3 | 0.405 | 0.508 | 0.609 | 0.709 | 0.815 | 0.914 | 0.935 | 0.957 | 0.978 |
| Generalized random forests (GRF) | 0.043 | 0.065 | 0.087 | 0.11 | 0.202 | 0.303 | 0.407 | 0.511 | 0.614 | 0.716 | 0.817 | 0.915 | 0.937 | 0.959 | 0.98 |
| Gradient boosting machines (GBM) | 0.052 | 0.077 | 0.1 | 0.124 | 0.215 | 0.313 | 0.414 | 0.511 | 0.605 | 0.704 | 0.804 | 0.901 | 0.925 | 0.949 | 0.974 |
| Light gradient boosting machines (LightGBM) | 0.058 | 0.084 | 0.107 | 0.133 | 0.222 | 0.322 | 0.415 | 0.51 | 0.6 | 0.692 | 0.788 | 0.887 | 0.913 | 0.936 | 0.961 |
| Quantile regression neural networks (QRNN) | 0.046 | 0.069 | 0.092 | 0.114 | 0.21 | 0.306 | 0.407 | 0.504 | 0.602 | 0.7 | 0.802 | 0.9 | 0.926 | 0.951 | 0.974 |
| Mean combiner | 0.047 | 0.069 | 0.091 | 0.114 | 0.209 | 0.309 | 0.414 | 0.515 | 0.612 | 0.712 | 0.813 | 0.906 | 0.932 | 0.954 | 0.976 |
| Median combiner | 0.047 | 0.069 | 0.09 | 0.113 | 0.206 | 0.306 | 0.41 | 0.511 | 0.609 | 0.711 | 0.812 | 0.906 | 0.93 | 0.953 | 0.976 |
| Best learner | 0.058 | 0.084 | 0.107 | 0.133 | 0.222 | 0.322 | 0.415 | 0.51 | 0.6 | 0.692 | 0.788 | 0.887 | 0.913 | 0.957 | 0.978 |
| Stacking with QR | 0.029 | 0.055 | 0.082 | 0.109 | 0.206 | 0.309 | 0.404 | 0.502 | 0.596 | 0.696 | 0.797 | 0.905 | 0.931 | 0.955 | 0.977 |
| Stacking with QRF | 0.051 | 0.075 | 0.1 | 0.125 | 0.218 | 0.319 | 0.412 | 0.509 | 0.6 | 0.698 | 0.796 | 0.903 | 0.927 | 0.951 | 0.976 |
| Stacking with GRF | 0.051 | 0.076 | 0.1 | 0.125 | 0.218 | 0.319 | 0.412 | 0.507 | 0.601 | 0.699 | 0.797 | 0.904 | 0.927 | 0.951 | 0.976 |
| Stacking with GBM | 0.027 | 0.055 | 0.081 | 0.103 | 0.2 | 0.305 | 0.401 | 0.502 | 0.599 | 0.696 | 0.796 | 0.902 | 0.927 | 0.95 | 0.975 |
| Stacking with LightGBM | 0.057 | 0.079 | 0.106 | 0.131 | 0.223 | 0.315 | 0.409 | 0.503 | 0.597 | 0.691 | 0.788 | 0.893 | 0.913 | 0.94 | 0.963 |
| Stacking with QRNN | 0.028 | 0.055 | 0.082 | 0.108 | 0.206 | 0.308 | 0.405 | 0.502 | 0.595 | 0.695 | 0.797 | 0.905 | 0.93 | 0.955 | 0.977 |

Figure 6. Sample coverage of the predictions of the algorithms at the various quantile levels. The closest the sample coverage to its nominal value (quantile level), the more reliable the predictions on average.

## 4.2 Importance of base learners in ensemble learning

Figure 7 presents the ranks of the predictions of the base learners at the various quantile levels based on the importance of these predictions as predictors in ensemble learning in the application of interest. According to explainable ML procedures of both the GRF and LightGBM algorithms, the predictions of LightGBM consist the most important predictor, while the predictions of QRF and GRF are also important. These results are in agreement with the ranks of the individual learners based on the quantile skill score (Figure 5b).

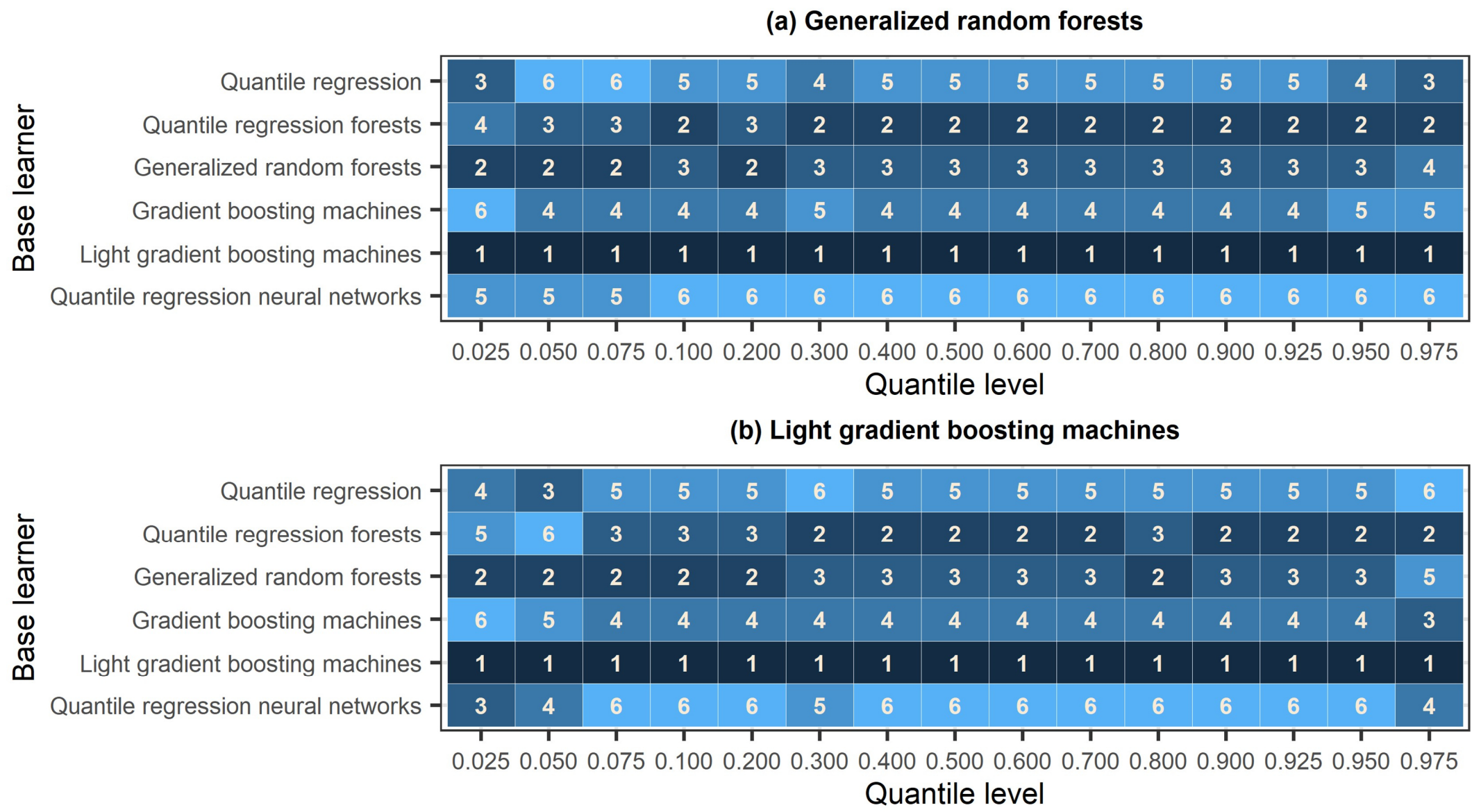

Figure 7. Ranking of the base learners at the various quantile levels based on (a) generalized random forests and (b) light gradient boosting machines. The smaller the rank and the darker the colour, the more useful the predictions of the base learners.

## 4.3 Importance of predictor variables

Figure 8 presents the order at each quantile level of the predictor variables based on their importance in uncertainty estimation in the application of this study. According to explainable ML procedures of both the GRF and LightGBM algorithms, the IMERG product offers more important predictors than the PERSIANN product, overall. Moreover, the station elevation appears in the second, third or fourth position for the quantile levels equal to or larger than 0.300 according to LightGBM. A final remark concerns the distance-based weighting made for producing the observations for the predictor variables. Because of this weighting, there should not be a priori expectations for the relative importance of the IMERG variables 1−4 (PERSIANN variables 1−4) and, indeed, the variable importance results confirm this in the sense that ordering of IMERG variables is not constant when varying the quantile level. In previous studies (Papacharalampous et al. 2023c; 2024), where unweighted satellite data were used as predictors, the closer grid data were consistently more important compared to more distant ones.

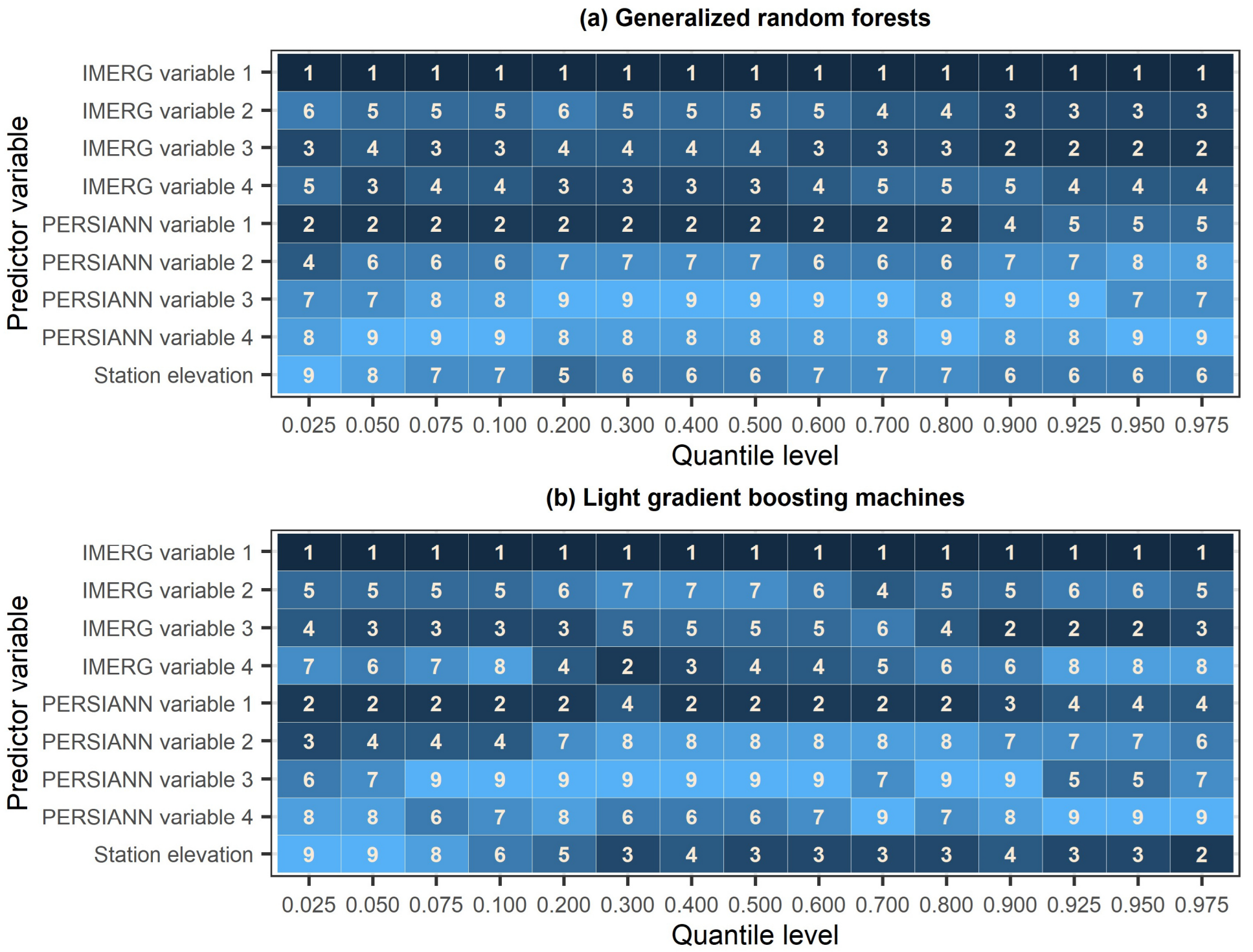

Figure 8. Ranking of the predictors at the various quantile levels based on (a) generalized random forests and (b) light gradient boosting machines. The smaller the rank and the darker the colour, the more important the predictor.

## 5. Discussion

As it is the case for all the categories of machine learning algorithms (Boulesteix et al. 2018), comparisons between ensemble learning methods and comparisons of such methods with individual machine learning algorithms should rely on large datasets. Furthermore, they should include as many algorithms as possible. Complying with these principles, the comparison conducted in this work is of large scale.

Overall, the central methodological contribution of this paper to the machine learning literature concerns the utilisation of quantile regression algorithms as combiners in ensemble learning methods for predicting the quantile. This new category of combiners can also be used for issuing predictions for the quantile through the combination of different machine and statistical learning algorithms (or even physics-based models, Tyralis and Papacharalampous 2021) for uncertainty estimation, even algorithms from families aside from the quantile regression one (see the review by Tyralis and Papacharalampous 2024). On the other hand, for cases in which the interest is in predicting other functionals that are measures of uncertainty as well (such as expectiles), algorithms that involve scoring functions which are strictly consistent for these functionals would be reasonable combiners.

Among the nine ensemble learners introduced in this work, ensemble learning of QR, QRF, GRF, GBM, LightGBM and QRNN (base learners) using QR as the combiner and ensemble learning of the same base learners using QRNN as the combiner were proven the best for the problem of uncertainty estimation while merging remote sensing and gauge-measured precipitation data at the monthly time scale. Indeed, the predictions of these ensemble learners scored better than the predictions of the other ensemble learners and the predictions of LightGBM, which is the best individual algorithm for this earth observation and geoinformation task (Papacharalampous et al. 2024). From a theoretical point of view, one could expect that an ensemble learning method outperforms individual algorithms (van der Laan 2007; Wolpert 1992). In this context, the selection of the combiner matters.

In our setting, the simplest combiner (linear QR) performed similarly to QRNN, even though QRNN might be expected to outperform QR. A possible explanation is that the ensemble learning step utilised a small number of predictor variables (i.e., base learners) and samples, limiting the ability to fully leverage machine learning's power. For example, QRNNs are known to improve generalization with more data. We expect that spatial settings with daily data (almost 30 times larger than monthly datasets) would enable better generalization of machine learning combiners. The relative performance of the nine new ensemble learners might differ in other uncertainty estimation problems, such as when predicting extremes (Tyralis and Papacharalampous 2023a; Tyralis et al. 2023). Therefore, all of them, and potentially others, should be evaluated on a problem-by-problem basis to identify optimal machine learning solutions.

## 6. Conclusions

### a. Methodological contributions

In this study, we formulated six ensemble learners based on stacking ideas and three simple ensemble learners for quantile prediction. These were created by combining six individual quantile regression algorithms in various ways, and constitute novel approaches introduced here for the first time in the machine learning literature.

### b. Contribution to remote sensing of precipitation

Beyond contributions to the machine learning field, the work also offers advancements in applied earth observation and geoinformation. Specifically, this study presents the first application of ensemble learning to estimate uncertainty while merging remote sensing and gauge-measured data, particularly for precipitation data. Furthermore, it proposes a novel feature engineering strategy for merging remote sensing and gauge-measured data. This strategy relies on distance-based weighting of satellite data and halves the number of satellite-based predictor variables with limited loss of information.

**c. Quantified results**

The six individual algorithms employed as base learners for all ensemble learners are quantile regression (QR), quantile regression forests, generalized random forests, gradient boosting machines, light gradient boosting machines (LightGBM), and quantile regression neural networks (QRNN). Each of these algorithms was also used to combine the base learners within one ensemble learning framework. The evaluation was based on quantile scores at multiple levels (0.025, 0.050, 0.075, 0.100, 0.200, 0.300, 0.400, 0.500, 0.600, 0.700, 0.800, 0.900, 0.925, 0.950, 0.975) of the predictive probability distribution.

For estimating uncertainty while merging remote sensing and gauge-measured data, ensemble learning using QR and ensemble learning using QRNN achieved the best performance. Compared to the QR reference method, these methods demonstrated performance improvements that range from 3.91% to 8.95% depending on the quantile level. LightGBM was the most effective individual base learner in this specific problem, providing performance improvements that range from 2.62% to 8.12%. Still, the ensemble learners significantly outperformed LightGBM at higher quantile levels. For example, the QRNN-based ensemble learning method demonstrated an improvement of 8.95% at the 0.975 level compared to an improvement of 5.85% by LightGBM. It is important to note that the relative performance of both the ensemble and base learners is likely to vary depending on the specific problem and should be evaluated on a case-by-case basis.

**Conflicts of interest:** There is no conflict of interest.

**Funding:** This work was conducted in the context of the research project BETTER RAIN (BEnefiTTing from machine lEarning algoRithms and concepts for correcting satellite RAINfall products). This research project was supported by the Hellenic Foundation for Research and Innovation (H.F.R.I.) under the “3rd Call for H.F.R.I. Research Projects to support Post-Doctoral Researchers” (Project Number: 7368).

**Acknowledgements:** We sincerely thank the Associate Editor and Reviewers, whose comments contributed to the improvement of the paper.

**Statement:** During the preparation of this work, the authors used Gemini in order to improve the readability and language of the manuscript. After using this tool, the authors reviewed and edited the content as needed and take full responsibility for the content of the published article.

## Appendix A Statistical software

The R programming language (R Core Team 2024) and the R packages listed in Table A1 were used to program the ensemble learners and conduct the application of this study.

Table A1. R packages used for conducting this study and their utilities.

| R package | Reference(s) | Utility in this study |
|---|---|---|
| caret | Kuhn (2023) | Data processing or visualization |
| data.table | Barrett et al. (2023) | |
| elevatr | Hollister (2023) | |
| ncdf4 | Pierce (2023) | |
| rgdal | Bivand et al. (2023) | |
| sf | Pebesma (2018, 2023) | |
| spdep | Bivand (2023), Bivand and Wong (2018), Bivand et al. (2013) | |
| tidyverse | Wickham et al. (2019), Wickham (2023) | |
| gbm | Greg and GBM Developers (2024) | Individual algorithm implementation |
| grf | Tibshirani and Athey (2023) | |
| lightgbm | Shi et al. (2024) | |
| qrnn | Cannon (2011, 2018, 2023) | |
| quantreg | Koenker (2023) | |
| scoringfunctions | Tyralis and Papacharalampous (2023b, 2024) | Scoring function computation |
| devtools | Wickham et al. (2022) | Report production |
| knitr | Xie (2014, 2015, 2023) | |
| rmarkdown | Allaire et al. (2023), Xie et al. (2018, 2020) | |